# A Channel Attention-Driven Hybrid CNN Framework for Paddy Leaf Disease Detection

Pandiyaraju V[1], Shravan Venkatraman[2], Abeshek A[3], Pavan Kumar S[4], Aravintakshan S A[5], Senthil Kumar A M[6], Kannan A[7]

[1,2,3,4,5,6,] School of Computer Science and Engineering, Vellore Institute of Technology, Chennai, Tamil Nadu, India

[7,] Department of Information Science and Technology, College of Engineering, Guindy, Anna University, Chennai, Tamil Nadu, India

pandiyaraju.v@vit.ac.in, shravan.venkatraman2021@vitstudent.ac.in, abeshek.a2021@vitstudent.ac.in, s.pavankumar2003@gmail.com , aravintakshan.sa2021@vitstudent.ac.in, senthil1185@gmail.com, akannan123@gmail.com

**ABSTRACT**

Farmers face various challenges when it comes to identifying diseases in rice leaves during their early stages of growth, which is a major reason for poor produce. Therefore, early and accurate disease identification is important in agriculture to avoid crop loss and improve cultivation. In this research, we propose a novel hybrid deep learning (DL) classifier designed by extending the Squeeze-and-Excitation network architecture with a channel attention mechanism and the Swish ReLU activation function. The channel attention mechanism in our proposed model identifies the most important feature channels required for classification during feature extraction and selection. The dying ReLU problem is mitigated by utilizing the Swish ReLU activation function, and the Squeeze-and-Excitation blocks improve information propagation and cross-channel interaction. Upon evaluation, our model achieved a high F1-score of 99.76% and an accuracy of 99.74%, surpassing the performance of existing models. These outcomes demonstrate the potential of state-of-the-art DL techniques in agriculture, contributing to the advancement of more efficient and reliable disease detection systems.

## 1. Introduction

Rice is one of the most essential foods in the world, consumed by about 36% of the world's population. India, contributing 20% to global rice production, is one of the leading producers of rice. Due to its high demand, farmers often use excessive amounts of synthetic fertilizers and pesticides, which has led to a decline in soil quality [1]. For efficient rice production, a balanced fertilizer with adequate quantities of various nutrients is necessary [2]. However, factors such as diseases, nutrient deficiencies, and pests can easily affect the quality and quantity of paddy yield [3]. Farmers' inability to detect leaf diseases and deficiencies early on often results in reduced rice yields. Factors like temperature, soil fertility, climatic conditions, and the presence of microorganisms such as bacteria and viruses play a crucial role in the growth of paddy plants. Diseases can affect various parts of the plants, such as leaves, tillers, and florets, manifesting as discoloration, lesions, or dryness [4]. Therefore, it is essential to detect these signs during the plant's growth. Identifying these diseases with the naked eye is challenging, necessitating the development of intelligent models to facilitate this process. Various automated models have been used to identify such leaf diseases [5].

Paddy leaf identification has become an important issue in recent years, as rice production is a significant economic and social challenge in many parts of the world. In South India, a majority of dishes are made from rice, making paddy production a critical requirement. The quantity of rice produced in the Tanjore District of Tamil Nadu, Andhra

Pradesh, and Punjab are major contribution to the food supply for the people of South and East India. However, diseases frequently attack paddy leaves, leading to a reduction in rice production. Therefore, plant health plays an important role in both Indian and global food production. In this context, the effective and efficient management of plant diseases is a significant research issue. Paddy leaf diseases must be identified at an early stage and prevented to ensure that the quantity of paddy and rice production is not affected. Leaf diseases can be analyzed effectively by collecting leaf images periodically and continuously monitoring them based on the results of the analysis. Most existing analytics on paddy leaf diseases use Machine Learning (ML) algorithms for effective image segmentation and classification. However, these methods require separate approaches for feature selection and classification and often provide a minimum level of required accuracy. In contrast, DL algorithms can more accurately extract and select contributing features. Therefore, DL classifiers provide the most accurate predictions regarding the development of paddy leaf diseases.

DL models are applied to identify many plant leaf diseases efficiently [6]. Various automated methods, such as using edges, clustering, borders, segmentation, thresholds, and active contours, are used to detect leaf diseases [7]. DL-based applications have been used for several plant disease identification and classification tasks [8, 9]. Convolutional Neural Networks (CNNs) are widely used in many real-time applications across various domains [10]. Many Artificial Intelligence (AI) techniques have been introduced and efficiently applied in agriculture, reducing the human effort required for tedious cultivation processes. Automated models using AI are applied to detect plant leaf diseases precisely [11].

This research proposes a novel hybrid CNN architecture for classifying paddy leaf diseases. Our model extends the SENet architecture with Squeeze and Excitation blocks using a channel attention mechanism and a Swish ReLU activation to facilitate efficient decision-making. Existing models discussed in the literature show drawbacks when it comes to handling vanishing gradients, and efficient color-based feature extraction. They also do not employ optimal image preprocessing techniques to prepare data for better model training. In our work, we create an image preprocessing pipeline consisting of a sequence of operations, specifically designed to enhance the quality of paddy leaf images and normalize variations in lighting and scale, thereby facilitating better training by our model. The Squeeze and Excitation block adjusts the weights of each feature map by considering the interdependencies between different feature channels in CNNs, thereby improving information propagation and cross-channel interaction. During feature extraction and selection, the channel attention mechanism is essential in determining which features are most significant for classification. Swish ReLU addresses the dying ReLU problem by introducing a smooth activation function that automatically adjusts its slope based on the input, preventing neurons from getting stuck at zero.

The remaining contents of the paper are organized as follows: The relevant literature is covered in Section 2, the proposed classifier model is explained in Section 3, the experimentation is described in Section 4, and the conclusion and the research's future directions are discussed in Section 5.

## 2. Related Works

Stephen et al. [12] proposed an approach to address the problem of rice leaf disease detection, which is traditionally performed manually by farmers and is prone to errors. Their research introduced a novel method utilizing an IBS-optimized DGAN and a 2D 3D CNN for feature extraction and for classification. The combination of 3D fast-learning blocks with 2DCNNs enhanced the detection of disease features, while the IBS algorithm improved the GAN's stability and prevented overfitting. The proposed method demonstrated superior performance with an accuracy of 98.7%, surpassing existing techniques like XGBoost and SVM. Overall, the research presented a robust solution for early and accurate rice disease detection, though it acknowledged the need for future studies to encompass a wider variety of plant diseases and features.

Chug et al. [13] proposed a framework to mitigate the significant economic losses caused by plant diseases through early detection. The authors developed an optimized hybrid DL model that combines pre-trained EfficientNet variants with ML classifiers using the Optuna framework. They collected a real-time image dataset of tomato early blight disease and achieved an accuracy, ranging from 87.55% to 100%, validated on additional datasets. The study highlighted the efficacy of their approach in reducing farmers' workloads and enabling timely disease treatment, though challenges like dataset biases and limited computational resources were noted. This work underscores the importance of integrating ML and DL for effective plant disease prediction.

Kumar et al. [14] came up with a solution for the inefficiency of traditional rice leaf disease identification methods, which rely on visual inspection and are prone to errors. They designed a multi-scale feature fusion-based RDTNet, which extracts features using local binary patterns, grayscale, and histograms of oriented gradients, combined with global and local features from transformers and convolutional blocks. This dual-module approach significantly improved classification accuracy, achieving 99.55% precision, 99.54% F1-score, and 99.53% accuracy. The study validated the model on various datasets, confirming its applicability for real-time rice disease diagnosis and potential use in monitoring other crops. The research presents a promising method for enhancing agricultural disease detection accuracy.

Bhagat et al. [15] employed a method for classifying leaf diseases by focusing on handcrafted features and ML classifiers, aiming to match the accuracy of DL models. They utilized a 3-level decomposition-based 2D-DWT for image feature extraction and PCA for dimensionality reduction. Stratified K-Fold validation ensured the maintenance of class ratios due to the small dataset size. The classification was performed using Random Forest and XGBoost, achieving accuracies between 97.73% and 100% across multiple datasets. The study demonstrated that handcrafted features and shallow classifiers could yield competitive results compared to DL models, providing an efficient alternative for leaf disease detection and classification in tomato, bell pepper, and potato species.

Sahu et al. [16] proposed a DD (Deep Dream) based disease detection architecture for crop leaves (CLDD) to address the problem of detecting crop leaves, which is crucial for preventing crop losses. The research introduces 24 Hybrid Deep Neural (HDN) models that integrate eight variants of EfficientNet (EffiNet B0-B7) for feature extraction with three ML algorithms: Stochastic Gradient Boosting, AdaBoost, and Random Forest. The DD technique is employed to segment and preprocess the lesions present in the leaves, enhancing interpretability and classification accuracy. The proposed models were evaluated on the PlantVillage tomato crop dataset, with the DD-EffiNet-B4-ADB model achieving the highest 96% accuracy. The study concludes that the proposed HDN models, especially when combined with DD segmentation, significantly improve the prediction accuracy of crop diseases, thus aiding farmers in early detection of diseases with a field performance of 100%.

Pandiyaraju et al. [17] developed a novel ensemble DL model to accurately classify tomato leaf diseases, addressing the limitations of existing ML-based classifiers in detecting new disease types. They present a weighted ensemble model that combines an enhanced weighted gradient optimizer (EWGO) with temporal constraints and an exponential moving average function into improved NASNet mobile and VGG-16 training techniques. Ten thousand photos of tomato leaves divided into nine disease groups make up the dataset that was used. The proposed model demonstrated superior performance with an accuracy of 98.7%, precision of 97.9%, and F1-score of 98.7%, outperforming existing models. The study concludes that the integration of DL CNN frameworks with EMA functions and gradient optimization significantly enhances the classification accuracy and provides an extensive approach for detecting tomato leaf diseases, with implications for improving crop yield and supporting sustainable agriculture.

Patil et al. [18] demonstrated an automated system for the classification of paddy leaf diseases in their early stage to support farmers in protecting their crops. The system employs ML algorithms such as AdaBoost and Bagging Classifier, along with a genetic algorithm (GA) and nearest neighbour algorithm (NNA) for disease identification. The images of paddy leaves undergo pre-processing steps including resizing, brightness correction, filtering, and geometric transformations to enhance image features. Feature extraction algorithms are then applied to create a relevant feature dataset, which is used by cascaded classifiers to improve accuracy. The proposed system, implemented using MATLAB, can be deployed on Android and Windows platforms, demonstrating the potential to assist farmers to detect diseases early, and crop protection with a high classification accuracy.

Jiang et al. [19] employed a method combining DL and support vector machine (SVM) technology to increase the accuracy of recognizing diseases in rice leaves. The research utilizes convolution neural networks (CNNs) to extract features from images of rice leaves, after which they are classified using support vector machines. The optimal SVM parameters were determined through 10-fold cross-validation, resulting in a recognition accuracy of 96.8%, which is higher than traditional backpropagation neural networks. The study concludes that the combination of SVM and DL provides an effective approach for crop disease diagnosis, offering higher accuracy and presenting a valuable method for future research in agricultural disease identification.

Al-Gaashani et al. [20] demonstrated a classification method to identify tomato leaf diseases leveraging transfer learning and feature concatenation to improve accuracy with limited training data. The research extracts features using NASNetMobile and MobileNetV2, which are then concatenated and reduced in dimension using principal component analysis. These are then fed into traditional ML classifiers, concluding that a multinomial logistic regression achieved the best performance with a 97% accuracy. The study concludes that the concatenation of features from different pre-trained models boosts classifier performance, making the method suitable for real-world applications where data is limited. The proposed approach offers a robust solution for accurate and early tomato disease diagnosis, with the potential for further development to identify new disease types and operate in diverse conditions.

Pantazi et al. [21] proposed an automated method for crop disease identification using Local One Class Classification and Binary Patterns (LBPs). Their approach involves dedicated One Class Classifiers (OCC) for each plant health condition, and the model was trained on vine leaves and tested on various crops, demonstrating high generalization capability. An original algorithm for conflict resolution between OCC was introduced, achieving a success rate of 95% across 46 plant-condition combinations. The study concludes that the proposed method offers a novel and effective approach to plant disease identification, with high generalization potential across different crop species, providing valuable support for farmers in early disease detection and crop protection.

Ramesh et al. [22] proposed a method for recognizing and classifying diseases in paddy leaves using the Jaya Algorithm on an Optimized Deep Neural Network (DNN_JOA). Preprocessing includes segmenting diseased areas using a clustering technique and transforming RGB images to HSV images for background removal. The DNN_JOA method classifies diseases by selecting the best weights with the JOA. Experimental results show high accuracy rates - Blast: 98.9%, Bacterial Blight: 95.78%, Sheath Rot: 92%, Brown Spot: 94%, and Normal Leaves: 90.57%. Compared with other classifiers like ANN, DAE, and DNN, DNN_JOA demonstrates superior accuracy and stability, achieving a testing accuracy of 97%. The study concludes that future improvements in recognition and classification methods can further enhance performance and reduce false classifications.

Haridasan et al. [23] presented an approach for classifying rice crop diseases to improve treatment efficiency and reduce yield losses. The system uses image processing, computer vision techniques, ML, and DL to identify diseases such as False smut, Rice blast, Sheath rot, Bacterial leaf blight, Brown leaf spot. After image pre-processing and segmentation, the diseases are classified using a combination of support vector machine classifiers and convolutional neural networks. The proposed DL-based strategy, utilizing ReLU and softmax functions, achieves a high validation accuracy of 0.9145. The system also recommends predictive remedies to assist in combating these diseases. This approach enhances the traditional methods of protecting paddy crops by providing accurate and efficient disease detection.

Yakkundimath et al. [24] explored the diagnosis of paddy blast disease utilizing three multi-layer CNN models for transfer learning: CapsNet, EfficientNet-B7, and ResNet-50. Based on the severity of the disease, field photos of blast disease that impact fifteen different types of paddy crops are categorized. With a dataset of 20,000 labeled photos, the study demonstrates that the CapsNet model delivers significant results with a validation efficiency of 93.29% and a testing efficiency of 90.79%. The average testing efficiencies of the ResNet-50 and EfficientNet-B7 models are 85.10% and 88.72%, respectively. The CapsNet model outperforms the other models in terms of classification and computational efficiency, demonstrating its effectiveness in diagnosing blast disease in paddy crops.

Sankareshwaran et al. [25] addressed the issue of detecting leaf diseases in rice plants using a novel approach - CAHA-AXRNet (crossover boosted artificial hummingbird algorithm-based AX-RetinaNet). This method aims to improve the accuracy and efficiency of detecting rice plant diseases, which are crucial for yield production and quality. The CAHA optimization model is used to tune the hyperparameters of the AX-RetinaNet model. To categorize rice plants as healthy or diseased, the study uses three datasets: rice plant, rice leaf, and rice disease datasets. The effectiveness of the technique is validated by performance indicators like accuracy, specificity, recall, precision, and F1-score. The CAHA-AXRNet approach outperforms other techniques for the detection of rice plant diseases, with an accuracy rate of 98.1%.

Bhimavarapu et al. [26] focus on detecting leaf diseases in rice leaves using DL to ensure sufficient food supply for a growing population. The study employs Convolutional Neural Networks (CNN) to analyze images of affected plant areas, extracting features to diagnose diseases efficiently. In order to reduce loss and improve prediction performance and classification accuracy, the suggested method adds an Optimizer Function and Improved Activation (IAOF) to the CNN model. This approach addresses the limitations of conventional manual identification methods, providing a faster, more effective, and cost-efficient solution. The IAOF-CNN model surpasses existing methods in output performance, demonstrating its effectiveness in rice disease prediction and classification.

## 3. Proposed Methodology:

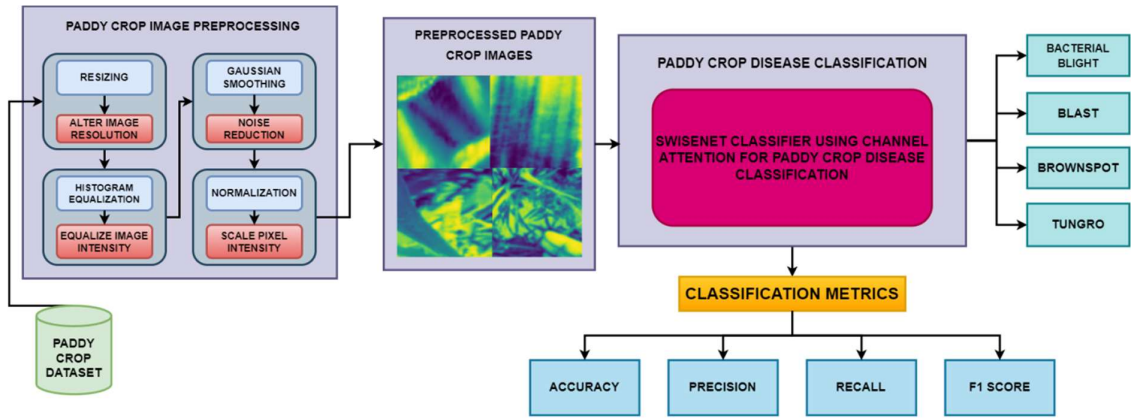

Figure 1. Overall Architecture of the Proposed Model

We procured a dataset containing 5932 images of rice leaves to classify four different types of paddy leaf diseases. Our preprocessing pipeline included techniques such as resizing, histogram equalization, Gaussian smoothing, and normalization. We applied histogram equalization to enhance the image contrast, resized the images to the necessary resolution, used Gaussian smoothing to reduce image noise, and normalized the pixel intensities for efficient computation.

After preprocessing, we fed the images into our classification model. We employed the proposed SwiSENet classifier, which utilizes channel attention for classification. Based on the features of the leaf images, we classified them into four leaf disease categories: Bacterial Blight, Blast, Brownspot, and Tungro. We evaluated the classification model using various performance metrics, including precision, F1-Score, recall, and accuracy. Figure 1 outlines the overall workflow of our research.

### 3.1 Dataset Exploration

We used a dataset developed by Sethy et al. [27], sourced from Rice Leaf Disease Image Samples available on Mendeley Data. This dataset comprises 5932 images, with sample images shown in Figure 2, depicting rice leaves affected by four different diseases: blast, bacterial blight, brown spot, and tungro. We implemented a 75-25 split to make use of this image dataset for our suggested approach, employing 75% of the images for model training and the remaining 25% for validation. As a result, we used 4449 of the 5932 photos for the training phase and the remaining 1483 images for validation.

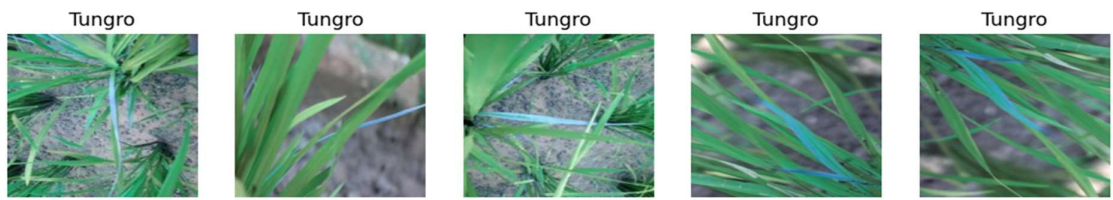

Figure 2.a Tungro Disease Images

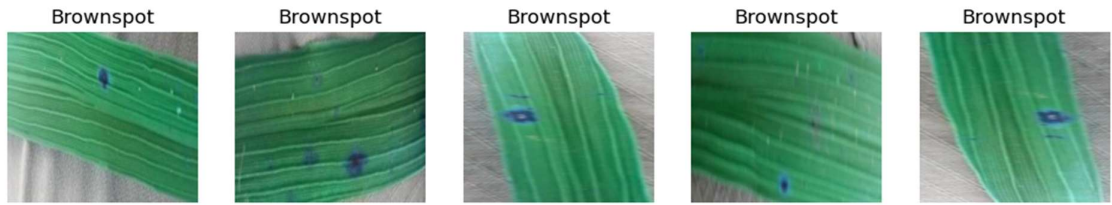

Figure 2.b Brown Spot Disease Images

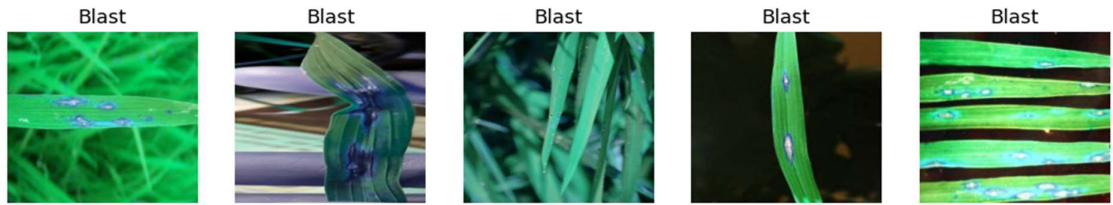

Figure 2.c Blast Disease Images

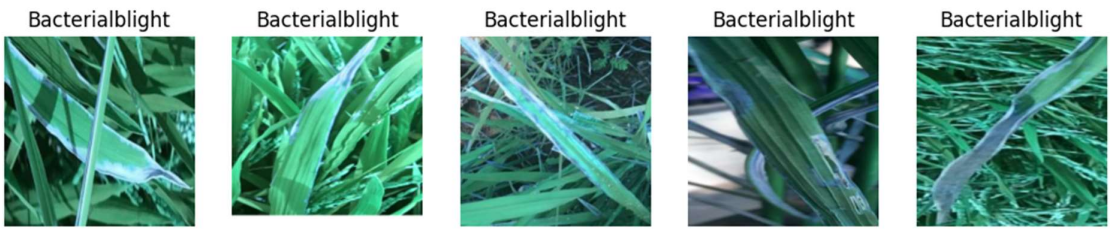

Figure 2.d Bacterial Blight Disease Images

The distribution of rice leaf data is illustrated in Figure 3. Among the 5932 images in the dataset, 27% represent brown spot disease, 24.3% depict blast disease, 26.7% are associated with bacterial blight disease, and the remaining 22% show leaves affected by tungro disease.

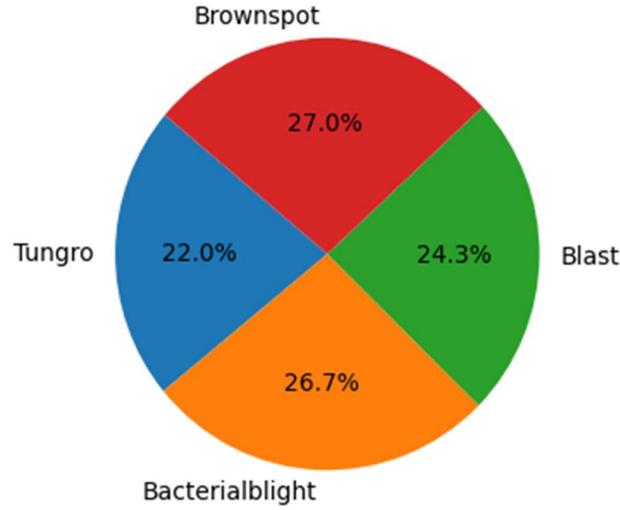

**Figure 3. Distribution of Images in the Dataset**

Table 1 displays the total number of images used in our experiment along with the number of images in each category.

**Table 1: Rice Leaf Disease Image Dataset Details**

| Class | Training Images | Validation Images |
|---|---|---|
| Tungro | 916 | 392 |
| Bacterial Blight | 1109 | 475 |
| Blast | 1008 | 432 |
| Brown spot | 1120 | 480 |
| **Total** | **4153** | **1779** |

*3.2 Preprocessing*

Before utilizing the dataset for classification, we preprocess the images to improve the proposed method's performance. Therefore, we carry out the following preprocessing techniques:

- Resizing
- Histogram Equalization
- Gaussian Smoothing
- Image Normalization

Initially, we apply image resizing to ensure uniform image sizes for the model. This is achieved by determining height and width scaling factors based on a target size and the original image dimensions. These scaling factors are then used to adjust the pixel coordinates x and y, resulting in the same image with different dimensions.

After resizing, we compute histograms representing the image intensities of each pixel. We then obtain the cumulative distribution function (CDF) based on the computed histogram using Equation 1.

$$CDF_i = \sum_{j=0}^{i} P_r(j) \quad (1)$$

Where $P_r(j)$ is the probability of occurrence of intensity level j.

With the cumulative distribution function (CDF) computed, we obtain the existing pixel intensities in the image and transform them using Equation 2.

$$I_{equalised}(x,y) = round(\frac{CDF(I(x,y)) - CDF_{min}}{CD_{max} - CDF_{min}} \cdot (L-1)) \quad (2)$$

Where, $I(x,y)$ and $I_{equalised}(x,y)$ are pixel intensities before and after histogram equalization respectively.

After equalizing the pixel intensities, we filter the Gaussian noise in the image using Gaussian smoothing, which we apply to all the pixels of the image. This smoothing operation involves a convolution operation between the image and the Gaussian kernel, which is generated for every pixel in the image based on Equation 3.

$$G(x,y) = \frac{1}{2\pi\sigma^2} \cdot e^{-\frac{x^2+y^2}{2\sigma^2}} \qquad (3)$$

Where G(x,y) is Gaussian kernel for the pixel (x,y) and $\sigma$ is the standard deviation of pixel intensities.

We normalize the Gaussian kernel before performing the smoothing operation. Once normalized, we carry out the smoothing operation using Equation 4.

$$I_{filtered} = \sum_m \sum_n I(x+m, y+n) \cdot G(m,n) \qquad (4)$$

Finally, we normalize the pixel intensities of the smoothed image to enhance performance and facilitate efficient computation, ensuring that the values fall within the range of 0 to 1. This is done using Equation 5.

$$I_{normalised}(x,y) = \frac{I(x,y) - minimum_{image}}{maxim\ value - minimum_{value}} \qquad (5)$$

The steps involved in the preprocessing are outlined in Algorithm 1. The results obtained after each step of preprocessing are shown in Figure 4.

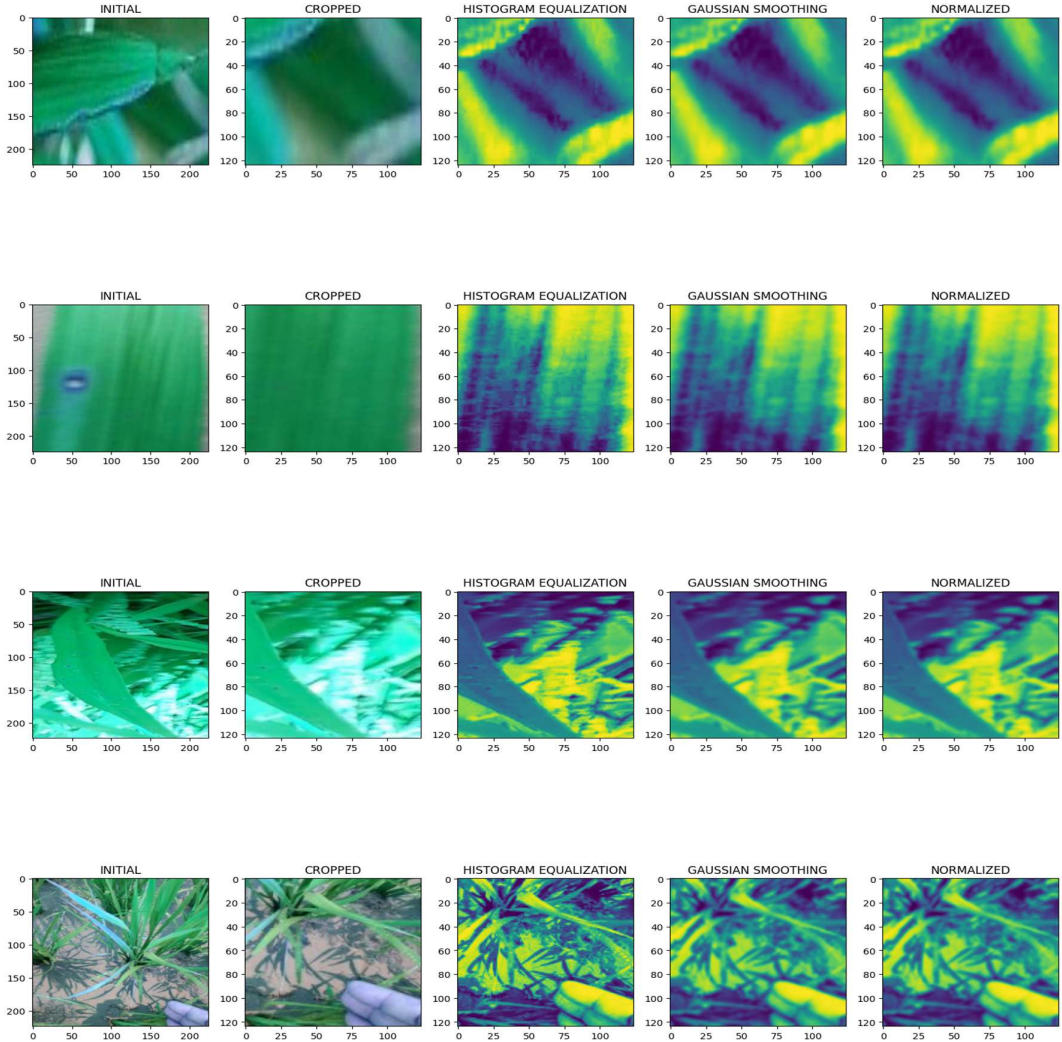

**Figure 4. Images Obtained After Pre-Processing**

## 3.3 Paddy Leaf Disease Classification

Figure 5 illustrates the layer architecture designed for constructing the proposed channel attention enabled hybrid CNN framework, named SwiSENet (Swish ReLU activated, SENet-inspired CNN). Preprocessed images are taken and sent to the classification model. These images move to the convolution block where various features are extracted. The max-pooling operation downsamples the input image by reducing the number of parameters. The feature map obtained after this pooling operation is used by the Convolution SE blocks (Conv_SE), which determine the number of features extracted from the feature map during each convolution operation.

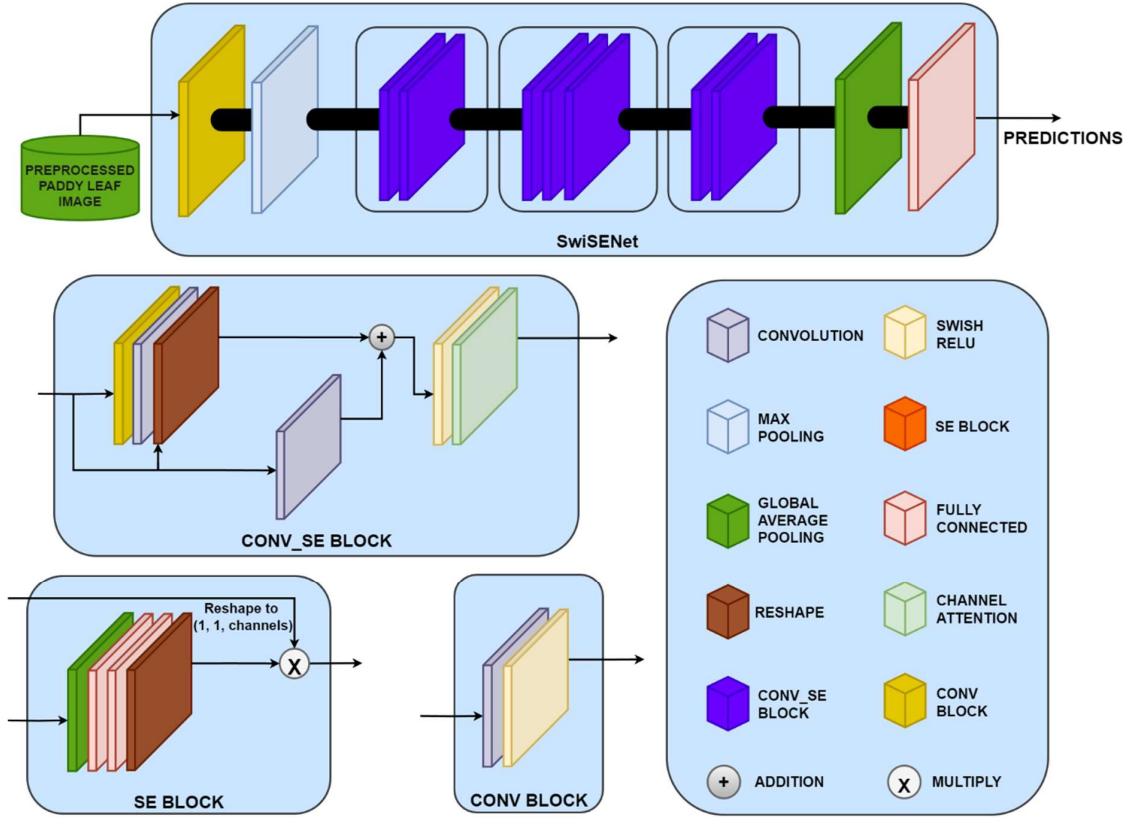

**Figure 5. Layer Architecture for Rice Leaf Disease Classifier using SwiSENet**

### 3.3.1 Conv_SE and SE blocks

The Convolution Squeeze and Excitation (Conv_SE) block extends the basic convolution block by incorporating additional components that enhance feature extraction. This block is pivotal in the network's architecture as it extracts the majority of features from the preprocessed paddy leaf images. The Conv_SE block consists of a convolution block, a Squeeze and Excitation (SE) block, a Channel attention layer, and the Swish ReLU activation function. The SE block within this architecture includes a global average pooling layer, two fully connected layers, and a reshape layer. The global average pooling layer performs a squeeze operation, reducing the spatial dimensions to a single value per channel. The two fully connected layers and the reshape layer execute the excitation operation, producing channel-wise weights that emphasize significant features, thereby improving the model's ability to focus on relevant information while reducing redundancy in the feature maps. This mechanism enhances the discriminative capability of our model.

Elaborating further, the SE block within the Conv_SE architecture functions as follows: the global average pooling layer condenses the spatial dimensions to a single value for each channel, effectively summarizing the information. This condensed representation is then passed through two fully connected layers and a reshape layer to generate channel-wise weights. These weights are applied to the feature maps, scaling the values to highlight the most critical features. This scaling operation ensures that the network prioritizes the most pertinent information, thereby enhancing the overall performance and efficiency of the feature extraction process.

### 3.3.2 Convolution Block

The Convolution block in our architecture is a fundamental unit designed to extract features from preprocessed paddy leaf images. It includes a convolution layer associated with the Swish ReLU activation function. The convolution layer employs kernels that slide across the input feature map with a specified stride value, performing convolution operations to extract features, which is mathematically expressed as:

$$O(a,b) = \sum_{m=-\infty}^{\infty} \sum_{n=-\infty}^{\infty} I(a-m, b-n) * K(m,n) \qquad (6)$$

Here, *O(x,y)* represents a value in the output feature map at position *(x, y)*, *I(x-i,y-j)* represents the pixel value in the input at position *(x-i,y-i)*, and *K(i,j)* represents a value of the kernel at position *(i,j)*.

The Swish ReLU function enhances the non-linearity of the network, aiding in the effective capture of complex patterns within the input data. By integrating the convolution operation with the Swish ReLU function, the block efficiently learns essential features from the input images, which are crucial for subsequent processing stages.

### 3.3.3 Channel Attention

The Channel Attention module, another key component of the Conv_SE block, aims to exploit the inter-channel relationships of the features to further refine the feature maps. This module includes an Average Pooling Layer, a Max Pooling Layer, and a Multi-Layer Perceptron (MLP). The feature map is processed separately through the Average Pooling Layer and the Max Pooling Layer, resulting in two distinct spatial context descriptors, $F_{avg}^c$ and $F_{max}^c$ respectively. These descriptors are then fed into a single MLP, producing two separate feature maps, which are subsequently aggregated using the addition operator to form a channel attention map. This map leverages the complementary information derived from the different spatial context descriptors. The resultant channel attention map, $M_c$, is then passed through a sigmoid activation layer, normalizing the weights and ensuring that the network effectively harnesses the most relevant features for classification tasks. The channel attention map function is expressed as follows:

$$M_c(F) = \sigma\left(W_1\left(W_0(F_{avg}^c)\right) + W_1\left(W_0(F_{max}^c)\right)\right) \qquad (7)$$

Where, $\sigma$ represents the sigmoid function, and $W_0$ and $W_1$ represent the MLP weights which are shared for both inputs [28].

### 3.3.4 SwisH-ReLU Activation Function

The Swish-ReLU activation function used in the Conv_SE block combines the benefits of the Swish activation function and the ReLU (Rectified Linear Unit) activation function. This composite function includes a tunable hyperparameter, $\alpha$, which controls the relative contributions of the outputs from Swish and ReLU. Through extensive experimentation, an optimal value of $\alpha = 0.5$ was determined, meaning that the activation function considers an equal contribution from both Swish and ReLU. This balanced approach effectively mitigates the Dying ReLU problem, where neurons become inactive, thereby maintaining a dynamic range of activations that enhance the learning process. ReLU activation function is expressed mathematically as

$$ReLU(k) = \max(0, k) \qquad (8)$$

Swish activation function can be defined as the product of the feature map with the sigmoid function along with a trainable parameter and is expressed as:

$$Swish(x) = x \cdot sigmoid(\beta x) = \frac{x}{1+e^{-\beta}} \quad (9)$$

Where $x$ is the feature map which is obtained from the convolution operation before activation and $\beta$ is a trainable parameter.

SwisH-ReLU is a combination of these two activation functions and is mathematically expressed as,

$$SwisH - ReLU(x) = \alpha * Swish(x) + (1-\alpha) * ReLU(x)$$

Table 2 outlines the structure of the proposed SwiSENet model, showcasing the interconnections and dimensional changes within its architecture.

**Table 2: Summary of proposed SwiSENet model illustrating connections between blocks and transformations in shape**

| Layer/Block | Output Shape | Param# |
| --- | --- | --- |
| ConvBlock_1 | (None,150,150,64) | 9,728 |
| MaxPooling_1 | (None,74,74,64) | 0 |
| ConvSEBlock_1 | (None,74,74,64) | 79,108 |
| ConvSEBlock_2 | (None,74,74,64) | 79,108 |
| ConvSEBlock_3 | (None,74,74,128) | 249,544 |
| ConvSEBlock_4 | (None,74,74,128) | 323,272 |
| ConvSEBlock_5 | (None,74,74,128) | 323,272 |
| ConvSEBlock_6 | (None,74,74,256) | 994,704 |
| ConvSEBlock_7 | (None,74,74,256) | 1,289,616 |
| GlobalAveragePooling_1 | (None,256) | 0 |
| Dense | (None,4) | 1028 |

**Workflow:**

In our SwiSENet model for classifying leaf diseases from paddy leaf images, the preprocessed images flow through various specialized blocks, each designed to extract and emphasize specific features crucial for accurate classification. This section outlines how the input preprocessed paddy leaf images are processed through the network, detailing the role and scientific rationale behind each block.

ConvBlock_1 initiates the feature extraction process, transforming the input into an output of shape (None, 150, 150, 64) using 64 filters. This initial convolution operation captures basic features like edges, textures, and simple patterns from the leaf images. The Swish ReLU activation function enhances this process by providing smooth non-linearity, aiding in better gradient flow and allowing the model to learn intricate patterns more effectively. Following the ConvBlock_1, we employ MaxPooling_1 to reduce the spatial dimensions of the feature maps while retaining the most prominent features. This step is crucial for managing computational efficiency without sacrificing critical information. It simulates the process of focusing on areas of interest on the leaf surface, ignoring less relevant details, thereby streamlining our attention towards potential disease indicators like discoloration or lesions.

The model then processes the features through a series of Conv_SE blocks. In ConvSEBlock_1 with an output shape of (None, 74, 74, 64), a convolution block, an SE block, a Channel attention layer, and the Swish ReLU activation function work together. The convolution block here further refines the feature extraction, emphasizing specific details such as disease spots and texture anomalies on the leaf surface. The SE block, comprising a global average pooling layer, two fully connected layers, and a reshape layer, performs a squeeze operation to condense spatial dimensions into a single value per channel. This operation reduces redundancy and emphasizes channel-wise significant features. The excitation operation then scales these features to highlight the most crucial aspects, ensuring that disease-specific features are more pronounced.

The transition to a higher filter count in ConvSEBlock_3 (None, 74, 74, 128) allows the model to capture more detailed patterns, such as subtle color variations and complex textures associated with different disease stages. Each Conv_SE block leverages the SE mechanism to prioritize important features dynamically, thereby improving the model's ability to discern between healthy and diseased regions. ConvSEBlock_4 and ConvSEBlock_5 (output shape: None, 74, 74, 128) enhances the model's capacity to extract intricate details, ensuring that the discriminative attributes of diseases such as nutrient deficiencies are more pronounced. The repeated application of SE and channel attention mechanisms ensures that each layer emphasizes the most relevant features, reducing redundancy and focusing on crucial patterns.

With ConvSEBlock_6 and ConvSEBlock_7 (output shape: None, 74, 74, 256), the model's depth and complexity increase, allowing it to capture high-level abstract features. These layers are critical for understanding the overall structure and distribution of disease patterns across the leaf surface. The increase in filter count to 256 enables the model to represent these complex patterns effectively. GlobalAveragePooling then reduces the spatial dimensions, resulting in an output shape of (None, 256). This layer of global average pooling reduces the feature map to a single vector of 256 dimensions, summarizing the learned features across the spatial dimensions. This condensation is crucial for feeding into the dense layer, ensuring that the model retains only the most significant information for classification. Finally, the Dense layer with an output shape of (None, 4) uses 1,028 parameters to perform the classification. This fully connected layer takes the condensed features and maps them to four classes, corresponding to the different types of leaf diseases being classified. Algorithm 2 maps out the workflow of our SwiSENet model, designed to classify various diseases in paddy leaves.

## 4. Results and Discussion

Our proposed model was trained and validated on a machine that features an Intel Xeon CPU running at 2.00GHz with support for KVM virtualization and an NVIDIA Tesla P100-PCIE-16GB GPU. Table 3 provides specifics on the parameters and corresponding coefficients that were utilized to train the model.

**Table 3: Specification of SwiSENet Parameters**

| Parameters | Coefficients |
|---|---|
| Total Number of layers | 180 |
| Total Trainable Parameters | 33,49,380 |
| Learning Rate | 0.00005 |
| Maximum number of iterations | 100 |
| Image Shape | (300, 300) |
| Batch Size | 32 |
| Total size of Population | 5932 |

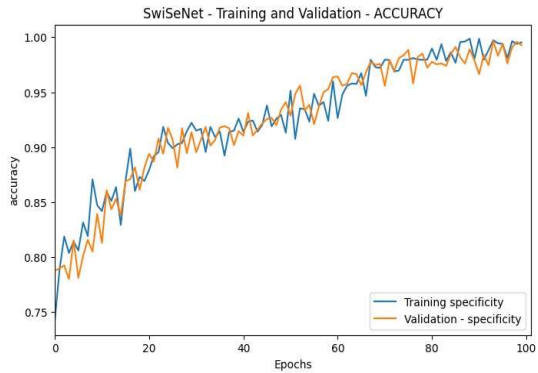
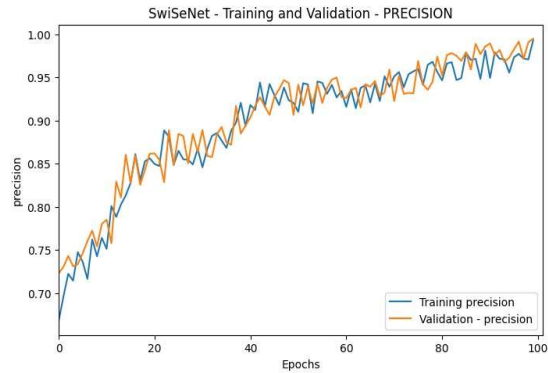

**Figure 6. Accuracy score obtained for each epoch during training and validation phases for the proposed SwiSENet classifier**

**Figure 7. Precision score obtained for each epoch during training and validation phases for the proposed SwiSENet classifier**

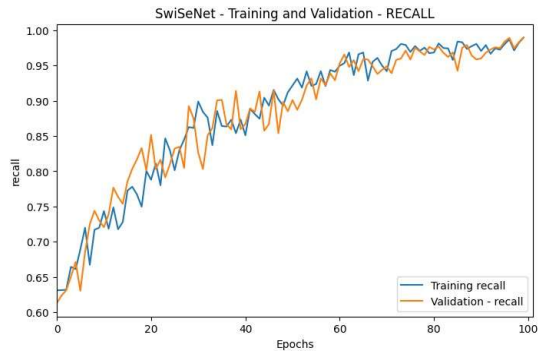
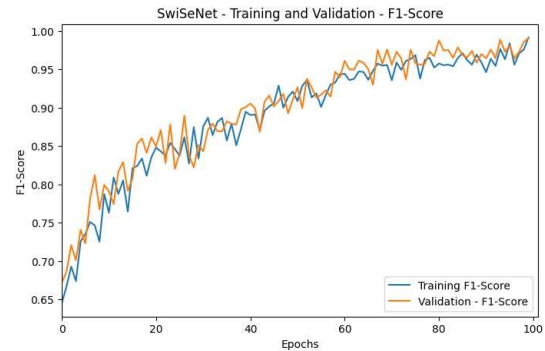

**Figure 8. Recall score obtained for each epoch during training and validation phases for the proposed SwiSENet classifier**

**Figure 9. F1 Score obtained for each epoch during training and validation phases for the proposed SwiSENet classifier**

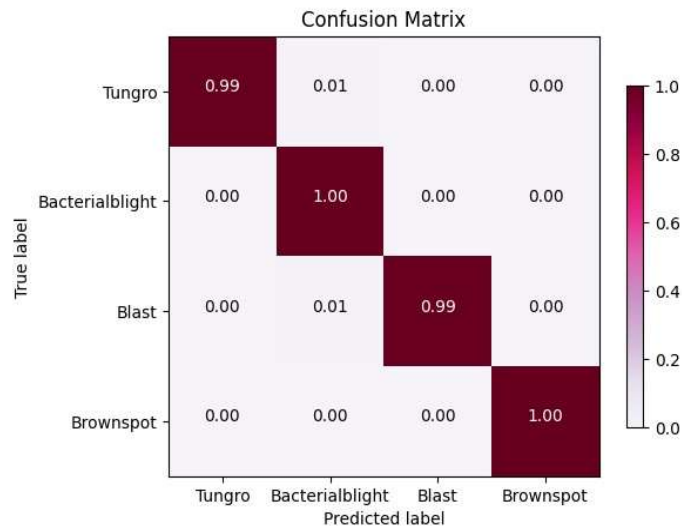

**Figure 10. Confusion Matrix obtained from the proposed SwiSENet classifier**

We evaluated our SwiSENet model's performance during training and validation by analyzing recall, accuracy, precision, and F1 scores.

The accuracy curves depicted in Figure 6, show that the model's accuracy consistently improved with each epoch, reaching 99.74% after 100 epochs. The close alignment of training and validation accuracy throughout the epochs indicates that the model performed uniformly well during testing. Precision curves illustrated in Figure 7, demonstrate a consistent upward trend in precision during both training and validation phases. By the end of the 100 epochs, the model achieved a precision of 99.8%, indicating its strong ability to make correct positive predictions reliably.

Figure 8 provides an analysis of recall on the benchmark dataset of paddy leaf images. The recall graph shows a gradual improvement, with the model attaining a recall rate of 99.7% after 100 epochs. This high recall rate underscores the model's effectiveness in minimizing false negatives, which is crucial for accurately detecting diseases. The F1-

score curves shown in Figure 9, reflect a steady rise in F1-score throughout the training and validation phases, culminating in a score of 99.7% after 100 epochs. This high score highlights the model's balanced performance in accurately detecting both positive and negative cases.

Finally, Figure 10 presents the normalized confusion matrix for the SwiSENet model's predictions on the four diseases under consideration. The normalized confusion matrix provides the proportions of correct and incorrect classifications relative to the total number of predictions for each class, allowing for a clearer comparison of model performance across different classes. The first row indicates that the model correctly identified 99% of tungro images, with only 1% misclassified as bacterial blight. The second and fourth rows show perfect identification of bacterial blight and brown spot images, with no errors. For blast disease images, the model misclassified 1% as bacterial blight, demonstrating its overall accuracy in disease classification.

The performance of our proposed work was compared with existing models, and this comparison is presented in Table 4.

Table 4: Performance comparison of proposed model with existing research

| Models | Performance Scores (in %) | | | |
|---|---|---|---|---|
| | Accuracy | Recall | Precision | F1 Score |
| CNN-SVM [29] | 98.43 | 88.82 | 85.0 | 86.83 |
| Fuzzy C-Means + Support Vector Machine [30] | 93 | 90.00 | 91.70 | 92.70 |
| Color feature + Support Vector Machine [31] | 90.55 | 94.70 | 92.10 | 90.60 |
| Histogram of Oriented Gradient + Support Vector Machine [32] | 93.76 | 90.50 | 95.60 | 88.30 |
| NasNet-Large [33] | 95.88 | 89.90 | 96.30 | 94.80 |
| DenseNet121 [34] | 97.44 | 92.10 | 93.80 | 89.10 |
| Inception-v3 [35] | 91.67 | 95.20 | 96.90 | 95.90 |
| ResNet50 [36] | 95.15 | 96.30 | 94.50 | 96.40 |
| Deep CNN [37] | 97.44 | 94.40 | 95.20 | 93.80 |
| K means clustering segmentation+ maykey optimization [38] | 98.10 | 96.20 | 97.60 | 96.80 |
| **Proposed SwiSENet** | **99.74** | **99.8** | **99.75** | **99.76** |

In our comparative study on paddy leaf disease classification, we analyzed various models across different methodologies, including traditional ML approaches and DL architectures. The CNN-SVM model demonstrated high accuracy at 98.43% but showed relatively lower precision and recall scores compared to its accuracy, indicating potential issues with false positives and negatives due to the complexity of distinguishing between similar classes within the dataset. The Fuzzy C-Means combined with Support Vector Machine achieved moderate results with an overall balanced performance across accuracy, precision, recall, and F1-score, suggesting a robust approach to handling the variability in the data. Models utilizing color features and Histogram of Oriented Gradients (HOG) combined with Support Vector Machines showed varied performances, with HOG performing slightly better in terms of recall but lower in precision, indicating a trade-off between capturing detailed texture information and maintaining class distinction.

NasNet-Large and DenseNet121 exhibited strong performances, particularly in terms of accuracy and F1-score, showcasing the effectiveness of convolutional neural networks (CNNs) in extracting complex features from images. Inception-v3 and ResNet50 further demonstrated the power of DL architectures, with ResNet50 achieving high precision and F1-score, indicating its superior ability to generalize across different classes without compromising on accuracy. The Deep CNN model mirrored the performance of DenseNet121, suggesting that while DL models generally outperform traditional ML approaches, the specific architecture can significantly impact the outcome. The K-means clustering segmentation combined with maykey optimization showed promising results, especially in terms of precision and recall, highlighting the potential of hybrid methods in image classification tasks.

Our proposed SwiSENet model excelled in all performance metrics, achieving an impressive accuracy of 99.74%. It demonstrated outstanding precision (99.8%), recall (99.75%), and F1-score (99.76%), showcasing its superior ability to classify paddy leaf diseases consistently and reliably. This exceptional performance is largely due to its unique architectural features and channel attention mechanism, which allow the model to dynamically adjust the importance of different channels in the feature maps. The integration of Swish activation functions and SENet-inspired channel attention mechanisms within SwiSENet's architecture is crucial to its effectiveness. These features enable the model to efficiently capture and utilize the most salient features from the input data, resulting in high consistency and reliability in its predictions.

## 5. Conclusions and Future Works

In summary, this research represents a significant advancement in plant disease management. By integrating state-of-the-art technology, meticulous data analysis, and image enhancement techniques, we have developed a channel attention enabled SENet inspired hybrid CNN framework with a swish ReLU activation function. The introduction of an attention mechanism, Squeeze and Excitation Blocks, and a modified activation function has enabled the proposed model to achieve superior performance, contributing to promising outcomes in leaf disease classification. In future, this work could be further enhanced by incorporating a swarm intelligence-based optimization algorithm, which could lead to faster learning and improved classification efficiency. This future enhancement holds the potential to make the SwiSENet model even more effective and applicable in real-world scenarios.

**Statements and Declarations:**

The "Ethical Responsibilities of Authors" declaration found in the "Instructions for Authors" has been read, comprehended, and complied with by all authors.

**Conflict of Interests:**

The authors declare no competing interests.

**Funding Information:**

No funding was received for this work.

**Data Availability and Access:**

The study's findings are supported by the Mendeley Rice Leaf Disease dataset, which can be accessed at https://data.mendeley.com/datasets/fwcj7stb8r/1.